\documentclass[10pt]{article}
\usepackage{pbox}
\usepackage{multirow}
\usepackage{setspace}
\usepackage{times}
\usepackage{graphicx}
\usepackage{natbib}
\usepackage{amsmath}
\usepackage{caption}
\usepackage[margin=1.0in]{geometry}
\begin{document}
\title{Next-Step Conditioned Deep Convolutional Neural Networks Improve Protein Secondary Structure Prediction} 

\author
{Akosua Busia$^{\dagger, *}$ and Navdeep Jaitly\\
\normalsize{Google Brain, Mountain View, CA, 94043, USA}\\
\\
\small{$^{\dagger}$ To whom correspondence should be addressed.}\\
\small{$^{\ast}$ Work completed as a member of the Google Brain Residency program (g.co/brainresidency)}\\
}

\date{}

\twocolumn[
\begin{@twocolumnfalse}
\maketitle

\newenvironment{sciabstract}{%
\begin{quote} }
{\end{quote}}

\begin{sciabstract}
\textbf{Motivation:} Recently developed deep learning techniques have significantly improved the accuracy of various speech and image recognition systems. In this paper we show how to adapt some of these techniques to create a novel chained convolutional architecture with next-step conditioning for improving performance on protein sequence prediction problems. We explore its value by demonstrating its ability to improve performance on eight-class secondary structure prediction.\\
\textbf{Results:} We first establish a state-of-the-art baseline by adapting recent advances in convolutional neural networks which were developed for vision tasks. This model achieves 70.0\% per amino acid accuracy on the CB513 benchmark dataset without use of standard performance-boosting techniques such as ensembling or multitask learning. We then improve upon this state-of-the-art result using a novel chained prediction approach which frames the secondary structure prediction as a next-step prediction problem. This sequential model achieves 70.3\% Q8 accuracy on CB513 with a single model; an ensemble of these models produces 71.4\% Q8 accuracy on the same test set, improving upon the previous overall state of the art for the eight-class secondary structure problem.\\
\textbf{Availability:} Our models are implemented using TensorFlow, an open-source machine learning software library available at \textit{TensorFlow.org}. We aim to release the code for these experiments as part of the TensorFlow repository. An early version of this work is available on \textit{arXiv.org}\\
\textbf{Contact:} \textit{apbusia@google.com}\\
\end{sciabstract}
  \end{@twocolumnfalse}
]

\section{Introduction}

\begin{figure*}[t]
\centerline{\includegraphics[width=0.9\textwidth]{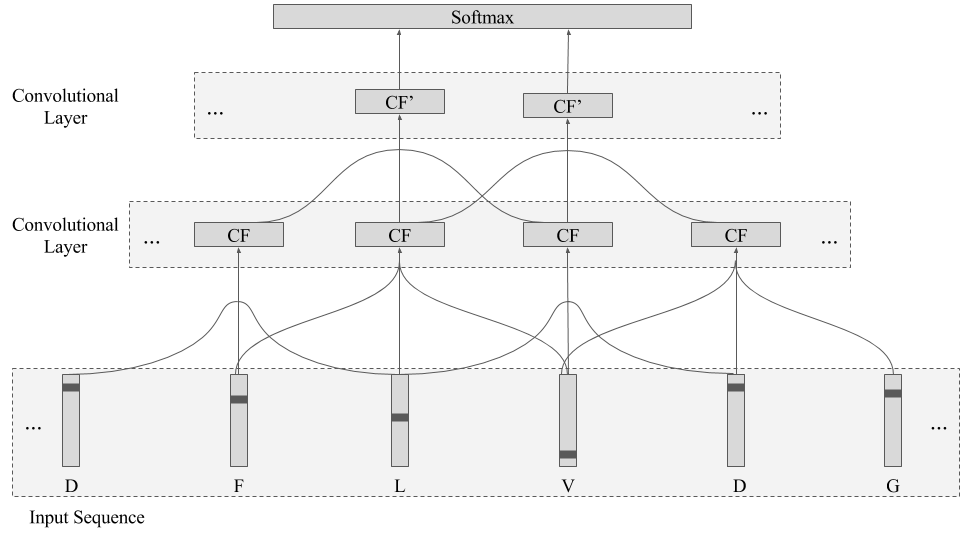}}
\caption{Illustration of convolutions acting on a segment of an example input sequence from the training data. The first convolutional layer consists of a single 3-filter, CF, which slides down the input sequence three vector-encoded residues at time. The second layer emphasizes the unique window-growth effect of successive convolutions: through stacked 3-filter convolutions, the fully-connected softmax layer at the top of the network effectively receives information from a total of five residues in the original input sequence.}
\label{fig:conv_layer}
\end{figure*}

Protein-based interactions are responsible for controlling a variety of vital functions: they are critical in driving the immune system, regulating breathing and oxygenation, controlling aging and energy usage, and determining drug response, with a protein's particular functional role determined by its structure (\citealp{1}; \citealp{2}). Over time, scientists have reached a consensus that a protein's structure primarily depends on its amino acid sequence--local and long-range interactions between amino acid residues and their side-chains are both a cause and consequence of protein secondary and tertiary structure \citep{3}. This hypothesis has driven a decades-long effort, spanning multiple disciplines, to deduce how protein sequence determines a protein's structural and functional properties (\citealp{3}; \citealp{4}).

As the number of proteins with known sequence continues to outpace the number of experimentally determined secondary and tertiary structures \citep{15}, computational approaches to protein structure prediction become increasingly desirable. Computational tools that can handle large amounts of data while making sufficiently accurate predictions of secondary structures can potentially serve to mitigate the cost and time burden in the experimental determination of protein structures.

Applications of machine learning to the protein secondary structure problem have a rich history: the use of neural networks for secondary structure prediction was pioneered by Qian and Sejnowski in 1988, and subsequent attempts using both recurrent (\citealp{6}; \citealp{7}) and convolutional (\citealp{7}; \citealp{8}; \citealp{9}; \citealp{10}) neural network architectures have managed to incrementally improve performance on the eight-class secondary structure problem. For example, by combining a deep convolutional neural network with a Conditional Random Field (CRF), Wang \textit{et al.} (2016) demonstrate convincing improvements from the use of consecutive convolutions and achieve 68.3\% per amino acid accuracy. In contrast, Li \& Yu (2016) used an ensemble of ten independently trained models, each comprised of a multi-scale convolutional layer followed by three stacked bidirectional recurrent layers, to reach 69.7\% per amino acid accuracy on the same test set, the highest accuracy previously reported.

We present two main contributions. First, we increase state-of-the-art performance for a single model on CB513 to 70.0\% by defining a new convolutional architecture for protein secondary structure prediction that integrates recent general purpose improvements in the field of deep learning. These include Batch Normalization \citep{11}, dropout and weight-norm constraint \citep{12}, residual connections \citep{13}, and multi-scale convolutional filters (\citealp{14}; \citealp{16}). Next, we introduce a new approach using chained models for next-step prediction of the target sequences of secondary structure labels (\citealp{17}; \citealp{18}; \citealp{chain}). We do this by incorporating next-step conditioning into our convolutional architecture: we condition individual predictions on both local sequence data and past target labels, and control for overfitting with scheduled sampling during training \citep{18}. With this technique, we further advance the above state-of-the-art result, achieving 70.3\% Q8 accuracy on CB513 with a single model; this outperforms the previous best result obtained by an ensemble of ten multitask recurrent-convolutional models \citep{7}.

\section{System and Methods}

\subsection{Data}

\renewcommand{\thefootnote}{\fnsymbol{footnote}}

We use two publicly available benchmark datasets, preprocessed by Zhou \& Troyanskaya (2014)\footnote{\textit{http://www.princeton.edu/~jzthree/datasets/ICML2014/}}: CullPDB and CB513. Each of these consists of protein sequences and structure label assignments downloaded from the Protein Data Bank archive (PDB) \citep{21}.

Each protein is represented as a sequence of amino acids, padded if necessary by no-sequence tokens to a sequence length of 700 residues. In turn, each amino acid is encoded as a 42-dimensional vector: the first twenty-one dimensions represent the one-hot encoding of the residue's identity (or the no-sequence token), while the remaining dimensions contain Position-Specific Scoring Matrices (PSSM) generated with PSI-BLAST (see Zhou \& Troyanskaya (2014) for more details) which we normalize via mean-centering and scaling by the standard deviation.

The full CullPDB dataset contains 6,128 proteins with less than 30\% sequence identity. Here, we use a subset of these data, which has been filtered to reduce sequence identity with the CB513 test data to at most 25\%. The resulting subset consists of 5,534 protein sequences--or equivalently 1,183,318 individual amino acid residues--for prediction. Consistent with S{\o}nderby \& Winther's (2014) arrangement, we randomly divide these 5,534 proteins into a training set of 5,278 proteins and a validation set of 256 proteins.

The 513 protein sequences in the CB513 dataset are used exclusively to measure test Q8 accuracy--that is, the percentage of the 84,765 amino acid residues for which the predicted secondary structure labels are correct.

\begin{figure*}[t]
\centerline{\includegraphics[width=0.8\textwidth]{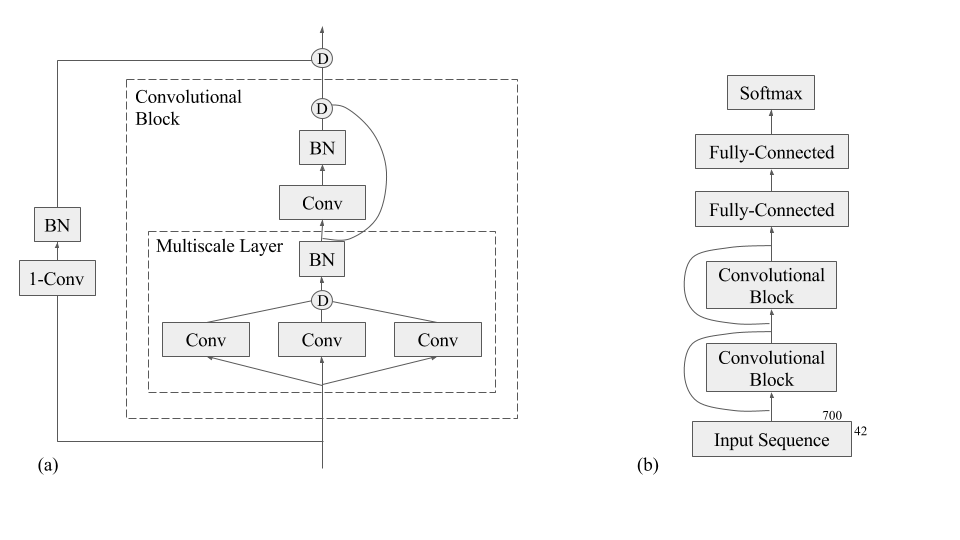}}
\caption{(a) A convolutional block composed of consecutive multi- and single-scale convolutions with Batch Normalization, rectified-linear activation (BN), and depth concatenation (D) of their outputs, plus modified residual connections between blocks. \ \ \ (b) The final convolutional architecture, composed of two convolutional blocks in the style of (a) followed by two fully-connected layers and a softmax output layer.}
\label{fig:conv}
\end{figure*}

\subsection{Architecture}

\subsubsection{Fully-Connected Model}

We begin with a simplistic baseline: we feed a fixed-sized context window of seventeen amino acids (padding edge-cases with no-sequence tokens) through five fully-connected layers of 455 rectified-linear units each in order to predict the secondary structure label of the central amino acid via a softmax output layer. The model is trained using a learning rate of 0.0004--which is reduced by 50\% every 35,000 training iterations.

We apply a combination of dropout and max-norm constraint on the model weights--this has been shown to give notable performance improvements even when the number of model parameters drastically outnumbers the size of the dataset \citep{12}. We follow a typical dropout approach: 20\% of the units in each layer (and their connections) are randomly dropped from the neural network during training, and at test time the the full network is used with the weights scaled by the dropout factor of 20\%. Max-norm contraint, on the other hand, rescales model weights as necessary to force the $l2$-norm of every unit's incoming weights to be at most a specified value \citep{12}; we find significant improvements from adding a weight-norm constraint of 0.04614.

\subsubsection{Convolutional Model}

\begin{figure*}[t]
\centerline{\includegraphics[width=0.65\textwidth]{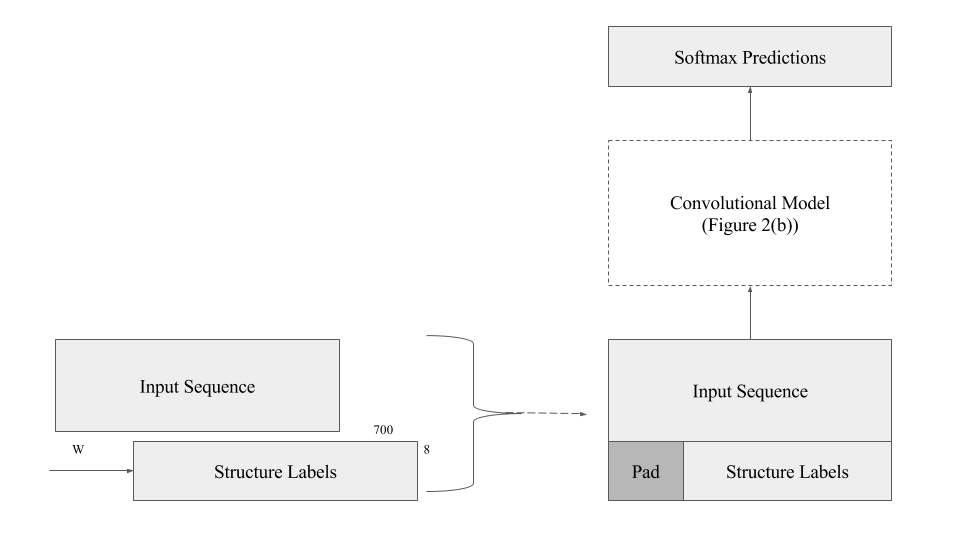}}
\caption{Demonstrates the introduction of next-step conditioning on previous labels by including a shifted copy of the input sequence's secondary structure labels as input to the convolutional model (see Figure \ref{fig:conv}(b)).}
\label{fig:cond}
\end{figure*}

Although a fully-connected layer--where every input of interest is connected to every neuron in the layer by a distinct weight--is usually considered the simplest type of layer in a neural network, in some senses a convolutional layer can be thought of as a further simplification. Whereas fully-connected layers can be understood as an attempt to capture information from the inputs simultaneously by defining a large number of neurons with distinct weights, convolutional layers consist of filters--smaller groups of neurons-- which look at segments of the input sequence at a time. Thus, a convolutional filter can be interpreted as sliding along the input sequence, reusing the same few weights on each local patch of the input.

Figure \ref{fig:conv_layer} offers an illustration of convolutions acting on a portion of an example protein sequence from the training set. The convolutional filters, CF and CF', are defined by their width--the number of inputs examined at a time--and depth--the number of neurons in the filter. In particular, the first convolutional layer in Figure \ref{fig:conv_layer} consists of a 3-filter which is repeated many times as it slides along the protein sequence. This visualization helps to make clear one of the major benefits of convolutions compared to a the previous fixed-window, fully-connected approach: in situations where local properties of the data are critical, stacking small filters--each of which examines a small local input patch at a time--introduces the ability to learn and maintain information about sequence dependencies at different scales. Specifically, Figure \ref{fig:conv_layer} shows how the filters in lower layers focus on extracting information from small local contexts--in this case, of three residues--while filters in higher layers cover correlations which are more spatially spread-out in the input sequence.

We develop a series of baseline convolutional models which exploit this property (see Table \ref{table:1}). We begin by adding a single convolutional layer to our fully-connected model, and add subsequently more complex convolutional structures to develop our architecture. We focus here on the particular details of our final convolutional architecture, which outperforms previously reported attempts at using convolutions for protein secondary structure prediction (\citealp{7}; \citealp{8}; \citealp{9}; \citealp{10}). Descriptions of the relevant improvements follow.

Batch Normalization is a technique for mitigating underfitting by normalizing layer inputs within the neural network architecture according to mini-batch statistics to reduce the shift in each layer's input distribution as the model trains \citep{11}. This has been shown to accelerate training on image recognition tasks, and we find that, for our problem, introducing batch normalization just prior to the application of a rectified-linear non-linearity followed by 40\% dropout provides a strong boost in validation accuracy. The fully-connected layers, however, instead maintain a max-norm constraint of 0.150 on the $l2$-norm of their weights.

Designed for image detection and classification tasks, the Inception architecture (\citealp{14}; \citealp{16}) makes heavy use of multi-scale convolutional layers which simultaneously apply different filter sizes to compute features on several scales. Similar to this Inception model, we use blocks composed of single- and multi-scale convolutional layers (see Figure \ref{fig:conv}(a)) as part of the larger neural network displayed in Figure \ref{fig:conv}(b). Table \ref{table:1} describes the number of filters, filter widths, and other architectural adaptations. However, we make several alterations to the typical Inception network which improve performance on the protein secondary structure problem.

We combine the approaches of three other successful image recognition models to develop a modified skip connection between layers of our model. The ResNET model \citep{13} improved performance on image classification by introducing residual connections consisting of additive identity shortcuts between the outputs of lower layers and the inputs to higher layers to improve information flow throughout the network. Huang, Liu, \& Weinberger's (2016) work on the DenseNet architecture provides an alternate approach to improving information flow which feeds the depth concatenation of the outputs of all previous layers in the model to any given layer. Finally, Network-in-Network architectures make use of 1x1-filter convolutions to control for high-dimensionality induced by successive multi-scale convolutions on images \citep{20}. Figure \ref{fig:conv} shows how we combine these aspects in our architecture. We break our model into blocks of multi- and single-scale convolutions, combining the outputs within blocks via depth concatenation. Then, similar to ResNet, we introduce connections between blocks, but find that depth concatenation of features gives vastly better training and validation performance than typical additive connections. We also make use of 1-filter convolutions to avoid an explosion in the number of features with the addition of blocks to the model. Thus, unlike previous image recognition models, our skip connections are formed by applying a 1-filter with depth 96 to condense the features learned by the layers in the previous block to a smaller fixed-number of features before concatenating those to the output of the current block (see Figure \ref{fig:conv}(a)).

Our final convolutional model in Figure \ref{fig:conv}(b) consists of two of these convolutional blocks connected by our modified skip connections, followed by two fully-connected layers of 455 units, the first of which receives a fixed-context window of eleven features produced by the final convolutional block. Each multi-scale layer contains 3-, 7-, and 9-filters of depth 64, and is followed by a single convolution with a filter size of 9 and a depth of 24. During training, we initialize the learning rate to 3.357$e^{-4}$, and reduce it by 60\% every 200,000 training iterations.

\subsubsection{Chained Model with Next-Step Conditioning}
\begin{table*}[t]
\centering
\small
\begin{tabular}{|c|c|c|c|c|c|c|c|}
\hline
\pbox{1.75cm}{\centering Multi-scale Convolution} & \pbox{1.75cm}{\centering Single-scale Convolution} & \pbox{1.25cm}{\centering \# Blocks} & \pbox{2.2cm}{\vspace*{3pt}\centering Window Size for Fully-Connected Layers\vspace*{2pt}} & \pbox{2.2cm}{\centering \# Fully-Connected Layers} & \pbox{1.6cm}{\centering Residual Connections} & \pbox{1.35cm}{\centering Validation Accuracy} & \pbox{1.35cm}{\centering CB513 Accuracy} \\
\hline
- & - & - & 17 & 5 & N & 71.4\% & 66.8\% \\
\hline
- & 7 x 32 & 1 & 17 & 5 & N & 71.8\% & 67.2\% \\
\hline
- & 7 x 32 & 2 & 17 & 5 & N & 72.2\% & 67.4\% \\
\hline
3 x 32 & \multirow{3}{*}{ - } & \multirow{3}{*}{1} & \multirow{3}{*}{11} & \multirow{3}{*}{5} & \multirow{3}{*}{N} & \multirow{3}{*}{72.5\%} & \multirow{3}{*}{67.4\%} \\
5 x 32 & & & & & & & \\
7 x 32 & & & & & & & \\
\hline
3 x 32 & \multirow{3}{*}{ - } & \multirow{3}{*}{1} & \multirow{3}{*}{11} & \multirow{3}{*}{2} & \multirow{3}{*}{N} & \multirow{3}{*}{72.8\%} & \multirow{3}{*}{68.1\%} \\
5 x 32 & & & & & & & \\
7 x 32 & & & & & & & \\
\hline
3 x 32 & \multirow{3}{*}{7 x 32} & \multirow{3}{*}{1} & \multirow{3}{*}{11} & \multirow{3}{*}{2} & \multirow{3}{*}{N} & \multirow{3}{*}{72.8\%} & \multirow{3}{*}{68.2\%} \\
5 x 32 & & & & & & & \\
7 x 32 & & & & & & & \\
\hline
3 x 64 & \multirow{3}{*}{9 x 24} & \multirow{3}{*}{2} & \multirow{3}{*}{11} & \multirow{3}{*}{2} & \multirow{3}{*}{N} & \multirow{3}{*}{74.3\%} & \multirow{3}{*}{69.0\%} \\
7 x 64 & & & & & & & \\
9 x 64 & & & & & & & \\
\hline
3 x 64 & \multirow{3}{*}{9 x 24} & \multirow{3}{*}{5} & \multirow{3}{*}{11} & \multirow{3}{*}{2} & \multirow{3}{*}{N} & \multirow{3}{*}{73.8\%} & \multirow{3}{*}{69.1\%} \\
7 x 64 & & & & & & & \\
9 x 64 & & & & & & & \\
\hline
3 x 64 & \multirow{3}{*}{9 x 24} & \multirow{3}{*}{2} & \multirow{3}{*}{11} & \multirow{3}{*}{2} & \multirow{3}{*}{Y} & \multirow{3}{*}{\textbf{74.8}\%} & \multirow{3}{*}{\textbf{70.0}\%} \\
7 x 64 & & & & & & & \\
9 x 64 & & & & & & & \\
\hline
\end{tabular}
\caption{Displays architectural details and Q8 accuracies for the series of baseline models. Performance on the held-out validation set is used to perform early stopping and select the step at which to report the CB513 test accuracies.}
\label{table:1}
\end{table*}
Here, we re-frame the protein structure prediction problem as a chained prediction task by introducing next-step conditioning, effectively using a language model over the sequences of structure labels. We model the dependencies between secondary structure labels by conditioning the current prediction on the previous structure labels in addition to the current input. This technique has led to substantial improvements in the performance of sequence-to-sequence models on complex tasks such as machine translation and pose prediction (\citealp{17}; \citealp{18}; \citealp{chain}).

Standard techniques for secondary structure prediction attempt to model the secondary structure ${\bf y} = y_1, y_2 \cdots y_L$ of a protein sequence of length $L$ using its input amino acid sequence ${\bf x}=x_1, x_2, \cdots x_L$, where $y_i$ is the secondary structure label of the amino acid at index $i$ and $x_i$ is the amino acid at index $i$ (or more generally, its descriptors, such as real-valued sequence information from PSSM matrices). This is typical of both convolutional and recurrent neural network approaches, which optimize a model under the following conditional distribution assumption:\vspace*{-5pt}
\begin{equation}
p\left(y_1, \cdots y_L | x_1 \cdots x_L\right) = p\left(y_1 | \bf{x}\right) \prod_{i=2}^L p\left(y_i | \bf{x}\right) \nonumber
\vspace*{-5pt}
\end{equation}
In contrast, introducing next-step conditioning forgoes the assumption of independence between predictions and allows us to instead attempt to model the probability distribution of $y_1 \cdots y_L$ using a chain rule decomposition:\vspace*{-5pt}
\begin{equation}
p\left(y_1, \cdots y_L | x_1 \cdots x_L\right) = p\left(y_1 | \bf{x}\right) \prod_{i=2}^L p\left(y_i | \bf{x}, y_{<i}\right)  \nonumber
\vspace*{-5pt}
\end{equation}

In typical applications of sequence-to-sequence modeling, this conditioning is implemented using recurrent neural network layers; here, we instead modify the approach to work as an extension of our convolutional architecture from the preceding section. We hold the convolutional architecture constant--with all hyperparameters kept as previously described--and introduce conditioning on a context of previous secondary structure labels during training by appending the input sequence (padded on the edges by no-sequence tokens) with a copy of the secondary structure labels shifted by the effective total window size $W$ of the convolutional model (see Figure \ref{fig:cond}). Note that $W$ depends on the architecture of the neural network and is a function of the widths of the convolutional filters and the window size that is input into the fully-connected layers.

Using this approach, for a given residue in the sequence, the model will integrate information not only from the relevant surrounding amino acids, but also from the additional context provided by the previous $W$ secondary structure labels. Thus, we effectively model $p\left(y_i | \bf{x}, y_{<i}\right)$ as $p\left(y_i | \bf{x}, y_{(i-W) \cdots (i-1)}\right)$ using our multi-scale convolutional neural network. 

During inference, the true previous secondary structure labels are not available. Instead, we find the label sequence ${\bf y}$ that maximizes $p\left(y_1, \cdots y_L | x_1 \cdots x_L\right)$ by decoding the input sequence using a standard left-to-right beam search. We use a beam search of size eight: for any given protein, we make sequential predictions and on each step maintain the eight most likely partial sequences of secondary structure assignments based on the running logarithmic probabilities produced by the model. These hypotheses are, in turn, fed as past context to the next beam search step. Additional details about this beam search procedure can be found in Sutskever, Vinyals, \& Le's (2014) work on sequence-to-sequence modeling with neural networks.

We find that the model is able to learn to achieve fairly high accuracy at predicting the next label when the previous labels are indeed correct; however, during inference the previous labels may not be entirely accurate as they are selected from the predicted probability distribution over the eight possible secondary structure classes during beam search. This mismatch between test and train conditions hurts the performance of the model. In light of this, we make use of a simple implementation of scheduled sampling to introduce uncertainty into the previous structure labels seen as context during training to make the model more robust to past mistakes.

In scheduled sampling, the model is increasingly fed its own predictions during training in place of the ground truth past labels \citep{18}. This means we sample the structure labels which are fed as past context during training, call them \textbf{c}, from the model's predictions at some rate $r$. Here, for a given protein, we obtain a sequence of sampled labels $\bf{\hat{y}}$ for each protein where $\hat{y}_i$ is drawn from the multinomial distribution described by $p\left(y_i | \bf{x}, \hat{y}_{(i-W) \cdots (i-1)}\right)$ instead of $p\left(y_i | \bf{x}, y_{(i-W) \cdots (i-1)}\right)$. Then, we update $\bf{c}$ according to \ \ \ \ \ \ \ \ \ \ \ \ \ \ \ \ \ \ \ \  \ \ \ $c_i = b \cdot \hat{y}_i + (1 - b) \cdot y_i$ where $b \sim \text{Binomial}(r)$. We initialize $r$ to 40\%, and gradually increase it by 10\% every 750,000 training iterations.

\subsection{Ensembling}

In addition to evaluating single models, we also train and evaluate an ensemble of ten independently trained models for each of the convolutional and next-step conditioned models. Each individual model is trained on a randomly sampled training and validation subsets. The outputs of these ten models are combined by averaging their predicted logarithmic probabilities over the secondary structure labels.

Let us denote the probability distribution predicted by the $m^{\text{th}}$ model in an ensemble as $p_m$. Then, a given prediction $\hat{y_i}$ of an ensemble of $N$ convolutional models is determined by evaluating\vspace*{-5pt}
\begin{equation}
\hat{p} = \frac{1}{N}\sum\limits_{m=1}^{N} \log p_m\left(y_i | \bf{x}\right) \nonumber
\vspace*{-5pt}
\end{equation}
and taking $\hat{y_i}$ to be the secondary structure label with the highest resulting probability.

For the next-step conditioned models, this ensembling process takes place sequentially during the beam search process. On the $k^{\text{th}}$ step of the search for a protein of length $L$, we calculate the average predicted logarithmic probabilities across the $N$ models in the ensemble based on the inputs $x$ and past context from the current hypotheses\vspace*{-5pt}
\begin{equation}
\hat{p} = \frac{1}{N}\sum\limits_{m=1}^{N} \log p_m\left(y_i | \bf{x}, \hat{y}_{(i-W) \cdots (i-1)}\right) \nonumber
\vspace*{-5pt}
\end{equation}
for each of the eight current partial hypotheses of secondary structure sequences. We add this value to the running logarithmic probability\vspace*{-5pt}
\begin{equation}
\sum\limits_{i=1}^{k-1} \log \hat{p}\left(y_i | \bf{x}, \hat{y}_{(i-W) \cdots (i-1)}\right) \nonumber
\vspace*{-5pt}
\end{equation}
and determine the new eight most probable partial hypotheses from this result. The ensemble's final predictions are obtained at the end of the beam search by selecting, out of the eight hypotheses, the sequence of secondary structure labels which has the highest probability according to the above equation at the final step $k=L$ of the beam search process.

We estimate error bars on the Q8 accuracy of these ensembles via bootstrapping: we train fifteen models total and calculate the Q8 performance metric for ten different randomly selected subsets of ten of these trained models, then estimate the standard error from the spread of the resulting values.

\subsection{Implementation}

We implement all of the architectures described in this paper using the open-source software library TensorFlow \citep{22}. During training, we use a randomly selected mini-batch of 50 protein sequences on each iteration, where a single protein sequence has an average length of roughly 214 amino acid residues. In all cases, biases in the model are initialized to 0.1; model weights are randomly initialized according to a zero-centered Gaussian distribution with a standard deviation of $3.0 / \sqrt{I}$ where $I$ is the number of inputs into the layer. We train these model parameters using TensorFlow's implementation of the cross-entropy loss (masking out padded sequence locations with zeros) and TensorFlow's ADAM optimizer \citep{19} with default beta and epsilon parameters. Finally, we use asynchronous distributed training across five machines with NVIDIA Tesla K20 GPU Accelerators and enforce early-stopping based on Q8 accuracy on the held-out validation set.

\section{Results}
\newcommand{\rpm}{\raisebox{.2ex}{$\scriptstyle\pm$}}

Our incremental study of convolutional architectures in Table \ref{table:1} demonstrates the impact of convolutions on Q8 accuracy: the introduction of a single convolutional filter to our simple fully-connected network increases Q8 accuracy on CB513 from 66.8\% to 67.2\%, an increase of roughly 0.4\%. We also demonstrate a measurable benefit of multi- versus single-scale convolutional filters: rows three and four of Table \ref{table:1} show that one multi-scale layer results in an accuracy boost equal to that of two single-scale convolutional layers before the same fully-connected network, despite the wider overall context window seen by the latter. However, a combination of single and multi-scale convolutional layers greatly outperforms either, increasing test accuracy by roughly 0.8\% regardless of a decrease in the size of the context window into the final fully-connected layers. Finally, we show that adding more convolutional blocks to the model improves performance by only 0.1\% whereas adding our modified residual connections to the model gives a larger performance boost of 1.0\%.

The addition of these residual connections brings the Q8 accuracy of our single trained convolutional model to $70.0 \pm 0.1\%$ on CB513, which is 0.6\% higher than 69.4\%, the previous state of the art by a single model on the same benchmark datasets \citep{7}. Furthermore, our next-step conditioned model boost performance on CB513 to $70.3 \pm 0.1\%$ Q8 accuracy, an additional improvement of 0.3\% which widens the gap over the previous state-of-the-art single model to 0.9\%. Note that unlike Li \& Yu's (2016) best model, we do not exploit multitask learning for boosted performance. Li \& Yu's (2016) model without multitasking achieves 68.8\% Q8 accuracy on CB513, so a truly direct comparison to our next-step conditioned model demonstrates a 1.5\% improvement. Moreover, this suggests our current results could be further improved by the addition of multitask learning.

While the performance of single models is instructive, the overall highest previously reported CB513 Q8 accuracy of 69.7\% was achieved by an ensemble of ten independently trained models \citep{7}. For fairer comparisons, we present the evaluations of ensembles of both our convolutional and next-step conditioned models in Table \ref{table:4}. An ensemble of ten of our convolutional and next-step conditioned models achieve $71.0 \pm 0.0\%$ and $71.4 \pm 0.1\%$ Q8 accuracy on the CB513 dataset, respectively. We, therefore, demonstrate an overall improvement in the state of the art of 1.7\%, again, without the use of multitask learning which was leveraged by the previous highest reported metric.
\begin{table}[t]
\centering
\small
\begin{tabular}{|c|c|c|}
\hline
Model & Multitasking & CB513 Accuracy\\
\hline
SC-GSN & Y & 66.4\% \\
Our Fully-Connected & N & 66.8\% \\
DCRNN with GRU & Y & 66.9\% \\
bLSTM & N & 67.4\% \\
DCRNN with bGRPU & N & 68.8\% \\
DeepCNF & N & 68.3\% \\
MUST-CNN & Y & 68.4\% \\
DCRNN with bGRU & Y & 69.4\% \\
Our CNN & N & 70.0\% \\
Our Conditioned CNN & N & \textbf{70.3}\% \\
\hline
\end{tabular}
\vspace{.1pt}
\caption{A comparison of the Q8 accuracy on CB513, which places the performances of current and recently published single machine learning models on the eight-class secondary structure problem in context.}
\label{table:2}
\end{table}
\begin{table}[t]
\small
\centering
\begin{tabular}{|c|c|c|}
\hline
Model & Multitasking & CB513 Accuracy\\
\hline
DCRNN & Y & 69.7\% \\
Our CNN & N & 71.0\% \\
Our Conditioned CNN & N & \textbf{71.4}\% \\
\hline
\end{tabular}
\vspace{.1pt}
\caption{A comparison of the Q8 accuracy on CB513 for ensembles of ten independently trained models.}
\label{table:4}
\end{table}
\begin{table*}[t]
\small
\centering
\begin{tabular}{c | c c c c | c c c c | c | c |}
\cline{2-11}
 & \multicolumn{8}{c|}{Single Model} & \multirow{3}{4em}{Frequency} &  \multirow{3}{5em}{Structure} \\
\cline{2-9}
 & \multicolumn{4}{c|}{Precision} & \multicolumn{4}{c|}{Recall} & & \\
\cline{2-9}
 & Our CNN & Our Cond & DCRNN & DeepCNF & Our CNN & Our Cond & DCRNN & DeepCNF & & \\
\hline
\multicolumn{1}{|c|}{Q8} & 0.700 & \textbf{0.703} & 0.694 & 0.683 & & &  & & & All\\
\multicolumn{1}{|c|}{H} & 0.840 & 0.841 & 0.836 & \textbf{0.849} & \textbf{0.936} & 0.932 & & 0.904 & \centering0.346 & $\alpha$-helix\\
\multicolumn{1}{|c|}{E} & 0.747 & \textbf{0.767} & 0.739 & 0.749 & \textbf{0.836} & 0.821 & & 0.833 & \centering0.217 & $\beta$-strand\\
\multicolumn{1}{|c|}{L} & \textbf{0.575} & 0.565 & 0.573 & 0.571 & 0.651 & \textbf{0.690} & & 0.657 & \centering0.192 & Coil\\
\multicolumn{1}{|c|}{T} & 0.571 & \textbf{0.577} & 0.549 & 0.530 & 0.524 & 0.524 & & \textbf{0.528} & \centering0.113 & Turn\\
\multicolumn{1}{|c|}{S} & 0.529 & \textbf{0.548} & 0.521 & 0.487 & 0.248 & 0.240 & & \textbf{0.255} & \centering0.082 & Bend\\
\multicolumn{1}{|c|}{G} & 0.454 & 0.487 & 0.432 & \textbf{0.490} & \textbf{0.309} & 0.285 & & 0.260 & \centering0.039 & $3_{10}$-helix\\
\multicolumn{1}{|c|}{B} & 0.656 & \textbf{0.676} & 0.558 & 0.433 & \textbf{0.050} & 0.041 & & 0.026 & \centering0.010 & $\beta$-bridge\\
\multicolumn{1}{|c|}{I} & & & & & & & & & \centering0.000 & $\pi$-helix\\
\hline
\end{tabular}

\vspace*{10pt}

\begin{tabular}{c | c c c | c c c | c | c |}
\cline{2-9}
 & \multicolumn{6}{c|}{Ensemble} & \multirow{3}{4em}{Frequency} &  \multirow{3}{5em}{Structure} \\
\cline{2-7}
 & \multicolumn{3}{c|}{Precision} & \multicolumn{3}{c|}{Recall} & & \\
\cline{2-7}
 & Our CNN & Our Cond & DCRNN & Our CNN & Our Cond & DCRNN & & \\
\hline
\multicolumn{1}{|c|}{Q8} & 0.710 & \textbf{0.714} & 0.697 & & & & & All\\
\multicolumn{1}{|c|}{H} & 0.841 & \textbf{0.846} & 0.832 & 0.936 & 0.936 & & \centering0.346 & $\alpha$-helix\\
\multicolumn{1}{|c|}{E} & 0.765 & \textbf{0.776} & 0.753 & \textbf{0.839} & 0.837 & & \centering0.217 & $\beta$-strand\\
\multicolumn{1}{|c|}{L} & \textbf{0.576} & 0.570 & 0.573 & 0.682 & \textbf{0.707} & & \centering0.192 & Coil\\
\multicolumn{1}{|c|}{T} & 0.585 & \textbf{0.591} & 0.559 & 0.537 & \textbf{0.542} & & \centering0.113 & Turn\\
\multicolumn{1}{|c|}{S} & 0.587 & \textbf{0.621} & 0.518 & \textbf{0.244} & 0.237 & & \centering0.082 & Bend\\
\multicolumn{1}{|c|}{G} & 0.499 & \textbf{0.528} & 0.429 & \textbf{0.308} & 0.290 & & \centering0.039 & $3_{10}$-helix\\
\multicolumn{1}{|c|}{B} & 0.696 & \textbf{0.786} & 0.554 & 0.047 & 0.047 & & \centering0.010 & $\beta$-bridge\\
\multicolumn{1}{|c|}{I} & & & & & & & \centering0.000 & $\pi$-helix\\
\hline
\end{tabular}
\caption{A comparison of precision and recall values for individual secondary structure classes on CB513. The frequency of each structure label in the training set and the name of each structure are also provided. Note that recall values were not reported for DCRNN on the CB513 dataset, and values for pi-helices are omitted as they do not appear at measurable frequency in the training set.}
\label{table:3}
\end{table*}

We also compute precision and recall values for our models, given in Table \ref{table:3}. Our convolutional model improves upon the precision values of DCRNN in all cases. Moreover, it improves over the precision values for DeepCNF \citep{10} for four out of the seven measurable classes, with the largest decrease relative to DeepCNF being only 3.6\%. Our convolutional model improves recall values for four classes and achieves comparable performance in all other classes, with the largest deficit to DeepCNF being only 0.7\%.

Interestingly, our convolutional and next-step conditioned models exhibit a kind of precision and recall trade-off: relative to the convolutional model's performance, the conditional model has higher precision for six out of the seven measurable structure classes but lower recall measures for five of these classes, with the widest margins of difference being 3.3\% and 2.4\%, respectively. However, the next-step conditioned model still improves recall in four classes when compared to DeepCNF, with its largest loss to DeepCNF being 1.5\%. Moreover, the conditional model improves precision values in six classes relative to DCRNN and four classes relative to DeepCNF, with the largest loss to these models being 0.8\% in both cases. Finally, it is interesting to note that the increases in performance by our two models are not relegated to the most common structure labels--with $alpha$-helices and $beta$-strands comprising more than half of the training set--but are mostly scattered throughout the rarest five secondary structure classes.

The precision and recall values for these two ensembles, also provided in Table \ref{table:3}, display a less pronounced trade-off. Although the ensemble of conditional models yields better precision metrics for six of the seven measurable structure classes, the recall metrics for the two ensembles are much closer than those for the single models; the convolutional ensemble is strictly better in terms of recall for only three classes, with the largest different being 1.8\% for the $3_{10}$-helix. The ensemble of next-step conditioned models improves upon the reported precision values for the DCRNN ensemble in six of the seven measurable classes, and this loss of 0.3\% is especially small compared to the largest improvement of 23.2\% in one of the rarest secondary structure classes. The ensemble of convolutional models, although generally slightly lower in precision than the next-step conditioned ensemble, actually outperforms the DCRNN ensemble for every class of secondary structure measurably represented in the training set.

\section{Discussion}

As demonstrated in Table \ref{table:1}, we build up to our state-of-the-art convolutional architecture in a principled way, making changes and additions which provide successive boost in performance. We start with fully-connected baseline to provide insight into the fundamental difficulty of the eight-class secondary structure problem, however, it also serves as an effective way to gauge the impact of our later architectural and algorithmic developments above and beyond the impact of effective regularization for neural networks. It is interesting to note that our fully-connected model outperforms SC-GSN's test metric of 66.4\% \citep{8} and nearly matches the 66.9\% Q8 accuracy achieved by Li \& Yu (2016) through a combination of multi-scale convolutions and a forward-passing recurrent neural network, despite being immensely more simplistic. This suggests that a large portion of the information relevant to predicting an amino acid's secondary structure arises from the local interactions amongst relatively few of the directly neighboring residues, with the recurrent layers in Li \& Yu (2016) model learning relatively little additional information from the remainder of the input sequence.

Convolutions improve upon a fixed-sized window method of sequence analysis by introducing the ability to learn and maintain information about sequence dependencies at different scales. Specifically, filters in lower layers focus on extracting information from local context, and filters in higher layers cover correlations which are more spatially spread-out in the input sequence. Multi-scale architectures take this one step further by applying different filter sizes simultaneously at each layer. Our results in Table \ref{table:1} align with the intuition that these convolutional filters help extract increasingly more helpful information from the input sequence. The last two rows of Table \ref{table:1} also highlight the boost in performance achieved by our specific architecture over naive applications of convolutional layers: the 1.0\% performance boost of adding our modified residual connections is much more impressive than the 0.1\% gained by naively adding more convolutional blocks to the model. We therefore believe that the development of these connections play a large role in our architecture's more accurate performance compared to past attempts to make use of convolutional filters (Tables \ref{table:2} and \ref{table:3}). Moreover, we believe these modified residual connections improve the model's ability to retain information about smaller local contexts within the sequence instead of allowing them to be washed out by longer-range features learned by subsequent convolutions over larger sequence patches. This, in addition to modeling information at multiple scales, may contribute to the improvements in precision and recall for less frequent classes in comparison to the reported values for DCRNN \citep{7} and DeepCNF \citep{10} in Table \ref{table:3}.

Despite its ability to capture sequence information at multiple scales in the input sequence, our convolutional architecture does not capture any interdependency between adjacent secondary structure labels because each label is predicted independently of the others. DeepCNF \citep{10} attempts to integrate information about the correlation between the structures of adjacent residues into a neural network; it does so using a CRF on the predicted structure labels. In contrast, we introduce a way to account for dependencies between output tokens by conditioning the current prediction of our multi-scale convolutional model on the previous structure labels in addition to the current input. In particular, we attempt to learn the distribution $p\left(y_i | \bf{x}, y_{(i-W) \cdots (i-1)}\right)$, which allows the model to learn extremely general distributions reflecting the impact of previous secondary structure labels on the current one and avoids the conditional independence assumptions amongst neighboring secondary structure labels which are required by other approaches such as CRF \citep{18}. In short, this allows our next-step conditioned model to learn richer distributions over our target sequences of secondary structure labels, making it a more enticing option for learning information from complex protein sequences.

Indeed, we find that adding conditioning on past secondary structure labels to the convolutional architecture has substantial impact on what the model learns. Our first attempt to train a next-step conditioned model--without scheduled sampling--was able to reach a Q8 accuracy of roughly 82\% on the validation set and 77\% on CB513 when the ground truth secondary structure labels were fed as past context, but under true inference with beam search these measures reduced to 71.9\% and 67.1\%, respectively. This large differential suggests that such an unstructured approach to conditioning on the past induces significant overfitting for protein secondary structure prediction; we posit that, without the relative robustness induced by scheduled sampling, this conditional model learns to almost always simply copy the previously seen label assignment during training, making it perform well on next-step prediction with ground truth labels but relatively poorly on inference when the true sequence of structure labels is not known. It is likely that the lack of assumptions regarding the nature of dependence between structure labels and the repetitive nature of secondary structure sequences combine to cause the model to ``over-learn'' dependencies between consecutive labels.

The use of scheduled sampling helps mitigate this: adding next-step conditioning to our convolutional model via a simple scheduled sampling approach improves its performance on the eight-class secondary structure problem by 0.3\%. We hypothesize, however, that some of this ``over-learning'' effect persists, and in particular that the recall metrics for individual classes in Table \ref{table:3}--especially those which are rare or tend to occur in shorter runs--suffer slightly from this copying tendency. We highlight here that the currently reported performance metrics for the next-step conditioned model are obtained without re-tuning of any architectural or regularization hyperparameters; it is likely that adjusting the amount of regularization in combination with more sophisticated sampling schedules will further widen the gap in performance between the convolutional and next-step conditioned models.

In the future, it may be more helpful to shift away from standard secondary structure prediction. There is a natural upper-bound on the maximum possible Q3 accuracy from inconsistencies in secondary structure assignment likely due to the coarseness and inherent arbitrariness of three-class labels \citep{Q3}. This is likely also the case for the eight-class instantiation of the problem. A better approach might be to predict the sequence of back-bone angles for the amino acids, since these are experimentally observed values.

\section{Conclusion}

The current work introduces a new technique for protein sequence prediction problems which combines the past successes using convolutional neural networks with language modeling approaches responsible for massive improvements in performance on many standard sequence-to-sequence tasks. Due to both its prominence and difficulty, we use the eight-class secondary structure problem to demonstrate the promise of this new approach, and in the processes we contribute two state-of-the-art results. First, by analyzing the impact of convolutional variants developed for recognition of natural images, we develop a multi-scale and residual convolutional architecture for secondary structure prediction which outperforms previous deep learning approaches on the same benchmark dataset with a single model and naive single-task objective. Then, by incorporating sequential modeling by adding next-step conditioning on structure labels, we further boost our newly achieved result. Thus, while there is much future work to be done to exploit conditioning by mitigating the overfitting introduced by next-step conditioning, this pushes state-of-the-art Q8 accuracy on the CB513 data upwards by 0.9\% relative to the highest previously reported result for a single model; by using ensembling, we demonstrate a total increase over the previous state of the art of 1.7\% by achieving 71.4\% Q8 accuracy on CB513. Finally, we note that, although we present architectures and techniques which we have honed specifically for performance on the secondary structure prediction problem, what we have presented here is a highly generalizable convolutional sequence model; this could be used for other protein sequence problems--such as solvent accessibility, contact number, or backbone angle predictions--as well as for unrelated sequence and time-series prediction problems.

\section*{Acknowledgements}

We wish to thank Jasmine Collins and Melody Guan, 2016 Google Brain Residents, for the valuable discussions regarding the datasets and project goals, as well as their assistance defining the initial baseline model. 

\vspace*{-12pt}

\begin{thebibliography}{}

\bibitem[Abadi {\it et~al}., 2016]{22}
Abadi, M., Agarwal, A., Barham, P., Brevdo, E., Chen, Z., Citro, C., ... \& Ghemawat, S. (2016). Tensorflow: Large-scale machine learning on heterogeneous distributed systems. arXiv preprint arXiv:1603.04467.

\bibitem[Anfinsen, 1977]{4}
Anfinsen, C.B. (1977). Studies on the principles that govern the folding of protein chains. \textit{Nobel Lectures in Molecular Biology: 1933-1975}, 401.

\bibitem[Bengio {\it et~al}., 2015]{18}
Bengio, S., Vinyals, O., Jaitly, N., \& Shazeer, N. (2015). Scheduled sampling for sequence prediction with recurrent neural networks. In \textit{Advances in Neural Information Processing Systems} (pp. 1171-1179).

\bibitem[Berman{,} Henrick{,} \& Nakamura, 2003]{21}
Berman, H.M., Henrick, K., \& Nakamura, H. (2003). Announcing the worldwide Protein Data Bank. \textit{Nature Structural Biology 10}(12): 980.

\bibitem[Breda {\it et~al}., 2006]{1}
Breda, A., Valadares, N.F., de Souza, O.N., \& Garratt, R.C. (2007). Protein structure, modelling and applications, {\it Bioinformatics in Tropical Disease Research: A Practical and Case-Study Approach}. National Center for Biotechnology Information, Bethesda (MD). Available from: http://www.ncbi.nlm.nih.gov/books/NBK6824/

\bibitem[Dill {\it et~al}., 2008]{3}
Dill, K.A., Ozkan, S.B., Shell, M.S., \& Weikl, T.R. (2008). The protein folding problem. \textit{Annual review of biophysics, 37}, 289.

\bibitem[Gkioxari{,} Toshev{,} \& Jaitly, 2016]{chain}
Gkioxari, G., Toshev, A., \& Jaitly, N. (2016). Chained predictions using convolutional neural networks. In \textit{European Conference on Computer Vision} (pp. 728-743). Springer International Publishing.

\bibitem[Guo{,} Ellrott{,} \& Xu, 2008]{2}
Guo, J.T., Ellrott, K., \& Xu, Y. (2008). A historical perspective of template-based protein structure prediction. In Zaki, M.J. \& Bystroff, C. (eds.), \textit{Protein Structure Prediction}, 2nd edn., Springer Science \& Business Media, Vol. 413, pp. 3-42.

\bibitem[He {\it et~al}., 2015]{13}
He, K., Zhang, X., Ren, S., \& Sun, J. (2015). Deep residual learning for image recognition. \textit{arXiv preprint arXiv:1512.03385.}

\bibitem[Huang{,} Liu{,} \& Weinberger, 2016]{15}
Huang, G., Liu, Z., \& Weinberger, K.Q. (2016). Densely connected convolutional networks. \textit{arXiv preprint arXiv:1608.06993.}

\bibitem[Ioffe \& Szegedy, 2015]{11}
Ioffe, S., \& Szegedy, C. (2015). Batch normalization: Accelerating deep network training by reducing internal covariate shift. \textit{arXiv preprint arXiv:1502.03167.}

\bibitem[Kihara, 2005]{Q3}
Kihara, D. (2005). The effect of long-range interactions on the secondary structure formation of proteins. \textit{Protein Science, 14}(8), 1955-1963.

\bibitem[Kingma \& Ba, 2014]{19}
Kingma, D., \& Ba, J. (2014). Adam: A method for stochastic optimization. \textit{arXiv preprint arXiv:1412.6980.}

\bibitem[Li \& Yu, 2016]{7}
Li, Z., \& Yu, Y. (2016). Protein secondary structure prediction using cascaded convolutional and recurrent neural networks. \textit{arXiv preprint arXiv:1604.07176.}

\bibitem[Lin{,} Chen{,} \& Yan, 2013]{20}
Lin, M., Chen, Q., \& Yan, S. (2013). Network in Network. \textit{arXiv preprint arXiv:1312.4400.}

\bibitem[Lin{,} Lanchantin{,} \& Qi, 2016]{9}
Lin, Z., Lanchantin, J., \& Qi, Y. (2016). MUST-CNN: A multilayer shift-and-stitch deep convolutional architecture for sequence-based protein structure prediction. \textit{arXiv preprint arXiv:1605.03004.}

\bibitem[Qian \& Sejnowski, 1988]{5}
Qian, N., \& Sejnowski, T.J. (1988). Predicting the secondary structure of globular proteins using neural network models. \textit{Journal of molecular biology, 202}(4), 865-884.

\bibitem[S{\o}nderby \& Winther, 2014]{6}
S{\o}nderby, S.K., \& Winther, O. (2014). Protein secondary structure prediction with long short term memory networks. \textit{arXiv preprint arXiv:1412.7828.}

\bibitem[Srivastava {\it et~al}., 2014]{12}
Srivastava, N., Hinton, G.E., Krizhevsky, A., Sutskever, I., \& Salakhutdinov, R. (2014). Dropout: A simple way to prevent neural networks from overfitting. \textit{Journal of Machine Learning Research, 15}(1), 1929-1958.

\bibitem[Sutskever{,} Vinyals{,} \& Le, 2014]{17}
Sutskever, I., Vinyals, O., \& Le, Q.V. (2014). Sequence to sequence learning with neural networks. In \textit{Advances in Neural Information Processing Systems} (pp. 3104-3112).

\bibitem[Szegedy{,} Ioffe{,} \& Vanhoucke, 2016]{16}
Szegedy, C., Ioffe, S., \& Vanhoucke, V. (2016). Inception-v4, inception-resnet and the impact of residual connections on learning. \textit{arXiv preprint arXiv:1602.07261.}

\bibitem[Szegedy {\it et~al}., 2015]{14}
Szegedy, C., Liu, W., Jia, Y., Sermanet, P., Reed, S., Anguelov, D., ... \& Rabinovich, A. (2015). Going deeper with convolutions. In \textit{Proceedings of the IEEE Conference on Computer Vision and Pattern Recognition} (pp. 1-9).

\bibitem[Wang {\it et~al}., 2016]{10}
Wang, S., Peng, J., Ma, J., \& Xu, J. (2016). Protein secondary structure prediction using deep convolutional neural fields. \textit{Scientific reports, 6.}

\bibitem[Zhou \& Troyanskaya, 2014]{8}
Zhou, J., \& Troyanskaya, O.G. (2014). Deep supervised and convolutional generative stochastic network for protein secondary structure prediction. In \textit{ICML} (pp. 745-753).

\end{thebibliography}
\end{document}